\documentclass[letterpaper, 10 pt, journal, twoside]{IEEEtran}

\IEEEoverridecommandlockouts

\usepackage[dvipsnames, table]{xcolor}
\usepackage[T1]{fontenc}
\usepackage{amsmath,amssymb}
\usepackage{scalerel}
\usepackage{siunitx}
\usepackage{comment}
\usepackage{graphicx}
\usepackage{subfigure}
\usepackage{multirow}
\usepackage[ruled,vlined]{algorithm2e}
\usepackage{bm}
\usepackage{url}
\usepackage{hyperref}
\usepackage{diagbox} 
\usepackage{soul}
\usepackage{balance}
\usepackage{tikz}

\usepackage{eso-pic}
\newcommand\AtPageUpperMyright[1]{\AtPageUpperLeft{%
 \put(\LenToUnit{1.8cm},\LenToUnit{-1cm}){%
     \parbox{\textwidth}{\raggedleft\fontsize{9}{11}\selectfont #1}}%
 }}%
\newcommand{\conf}[1]{%
\AddToShipoutPictureBG*{%
\AtPageUpperMyright{#1}
}
}

\newcommand\changes[1]{\textcolor{black}{#1}}
\input{settings/def_macros.tex}

\begin{document}

\title{\LARGE \bf D-VAT: End-to-End Visual Active Tracking \\ for Micro Aerial Vehicles}

\author{Alberto~Dionigi, Simone~Felicioni, Mirko~Leomanni, and Gabriele~Costante\thanks{ The authors are with the Department of Engineering, University of Perugia, 06125 Perugia, Italy {\tt\footnotesize \{alberto.dionigi, simone.felicioni, mirko.leomanni, gabriele.costante\}@unipg.it}.}
	\thanks{Code Repository: \url{https://github.com/isarlab-department-engineering/d-vat}}
}

\maketitle
\conf{\centering This work has been accepted to the IEEE Robotics and Automation Letters (RA-L). This is an archival version of our paper. Please cite the published version DOI: \url{https://doi.org/10.1109/LRA.2024.3385700}}

\begin{abstract}
Visual active tracking is a growing research topic in robotics due to its key role in applications such as human assistance, disaster recovery, and surveillance. In contrast to passive tracking, active tracking approaches combine vision and control capabilities to detect and actively track the target. Most of the work in this area focuses on ground robots, while the very few contributions on aerial platforms still pose important design constraints that limit their applicability. 
To overcome these limitations, in this paper we propose D-VAT, a novel end-to-end visual active tracking methodology based on deep reinforcement learning that is tailored to micro aerial vehicle platforms. The D-VAT agent computes the vehicle thrust and angular velocity commands needed to track the target by directly processing monocular camera measurements. We show that the proposed approach allows for precise and collision-free tracking operations, outperforming different state-of-the-art baselines on simulated environments which differ significantly from those encountered during training. \changes{Moreover, we  demonstrate a smooth real-world transition to a quadrotor platform with mixed-reality}.
\end{abstract}

\IEEEpeerreviewmaketitle
\section{Introduction} \label{sec:intro}

Micro aerial vehicles (MAVs) are gaining increasing interest thanks to their agility and low cost, which make them suitable for a wide variety of robotic tasks, especially those performed in cluttered or dangerous environments. Applications include transportation, exploration, surveillance, and tracking~\cite{emran2018reviewquadrotor}. 
In this paper, we focus on the visual active tracking (VAT) task, which requires a  \textit{tracker} vehicle to maintain visual contact with a dynamic  \textit{target}. {In contrast to {passive} tracking, where the pose of the camera is fixed, {active} tracking approaches actively regulate the camera pose by suitably controlling the vehicle, in order to keep the target inside the camera field-of-view (FoV). The VAT problem is far more challenging than passive tracking as it requires to directly map high-dimensional image data into suitable control actions}. Previous research on this problem combined a dedicated perception module (\eg an object detector) with a separate closed-loop control module for the \changes{vehicle motion~\cite{murray1994motion,das2018stable}}. This approach has two fundamental limitations: (i) the two modules are designed separately and not jointly optimized; (ii) their combination requires extra effort for tuning and implementation. 

A viable alternative to overcome these drawbacks is to adopt end-to-end deep reinforcement learning (DRL), which has already shown impressive results in many fields of \changes{robotics~\cite{berner2019dota, chaplot2020neural,devo2020deep}}. Recently, this paradigm has been explored for VAT~\cite{devo2021enhancing,dionigi2022vat}. Most of the related works focus on ground robots and take advantage of the physical characteristics of these platforms (\ie low dimensionality of the configuration space and limited number of possible actions) to facilitate the design of VAT policies.
However, much less attention has been devoted to more complex platforms such as MAVs, which require a more sophisticated policy to be learned by the DRL agent.
State-of-the-art (SotA) works have addressed this issue by relying on some simplifying assumptions, \eg by ignoring the vehicle dynamics \cite{xi2021anti} or by constraining the possible control actions to a predefined subset of the 
action space \cite{zhao2021active}.
Solutions based on these simplifications are, in general, less robust and performing. 

\begin{figure}[t]
    \centering
    \includegraphics[width=\linewidth]{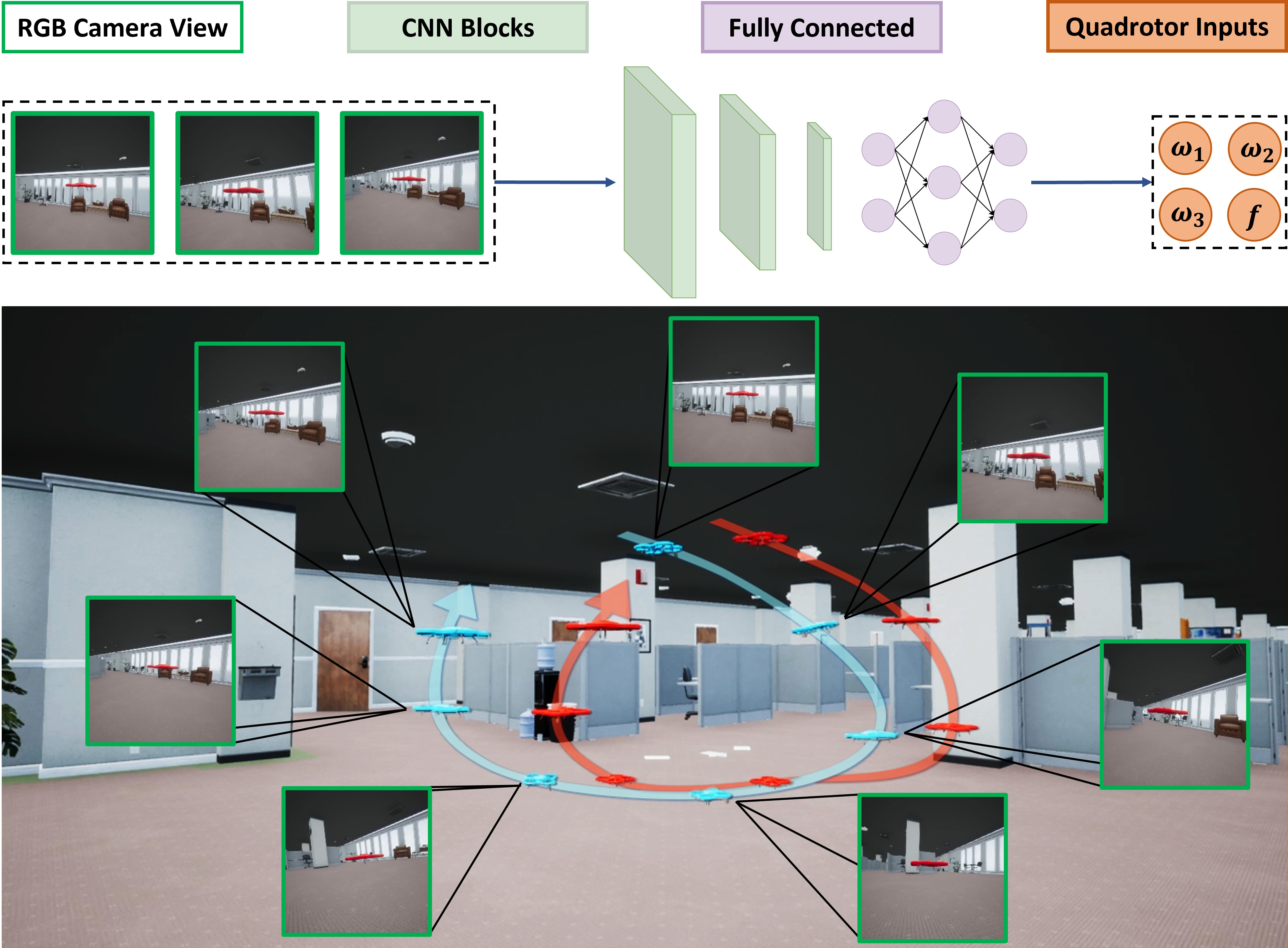}
    \caption{Overview of the VAT task. The \textit{tracker} MAV (blue) adjusts its position and orientation so as to keep the \textit{target} MAV (red) at the center of the camera FoV and at a predefined distance. Our approach exploits an end-to-end DRL-based VAT method that directly maps RGB images into thrust and angular velocity commands that are fed to the tracker.}
    \label{fig:overview}
\end{figure}

In this paper, we aim to remove these assumptions and  propose D-VAT, a novel end-to-end DRL-based continuous control model for visual active tracking that is tailored to MAV systems. D-VAT relies on a monocular setup, \ie it requires only an RGB image stream collected by an onboard camera to directly compute the thrust and angular velocity commands needed to track the target with high accuracy (see 
\cite{kaufmann2022bench} for a justification of such commands).
To the best of our knowledge, this is the first end-to-end approach that solves the VAT problem for MAVs without severely constraining the motion of the target or the tracker vehicle. We compare D-VAT to both model-based and data-driven SotA strategies on photorealistic simulated environments considerably different from those employed during training, where it achieves a much better tracking performance than these methods. \changes{Furthermore, we directly deploy D-VAT on a real drone without any fine-tuning, by employing a Mixed-Reality framework.}

The rest of this work is organized as follows: Section \ref{sec:related_work} contains literature review and details the paper contribution; Section \ref{sec:preliminary} provides the preliminary definitions; Section \ref{sec:approach} formalizes the considered tracking problem; Section \ref{sec:experiments} describes the experiments and discusses the results; Section \ref{sec:conclusions} draws the conclusions and outlines future research directions.

\section{Related Work}\label{sec:related_work}

In recent years, VAT has become a central research topic in robotics. VAT applications consider either pan-tilt-zoom (PTZ) vision sensors attached to a fixed base or cameras mounted on robotic vehicles to meet the goal of keeping the tracked object in sight. For instance, \cite{bellotto2012cognitive} presents a visual tracking solution that enables a PTZ camera to track the behavior of a moving person in surveillance applications. Along the same line, \cite{ribaric2004real} proposes a two layer architecture for real-time human motion tracking. In the context of mobile robots, VAT takes advantage of the control degrees of freedom of the vehicle to maintain the visibility of the tracked object. Most of the related approaches employ modular architectures that combine passive perception and motion control components\changes{~\cite{murray1994motion, das2018stable}}. In particular, ~\cite{hong2018virtual} couples the perception module with a low-level controller based on DRL. The former computes semantic segmentation maps from RGB images to obtain an intermediate representation that facilitates the agent in controlling the vehicle. 
Despite the significant results achieved by modular approaches such as the above ones, the combination of perception and control components poses, in general, important challenges. First, the modules are designed independently and not jointly optimized, reducing the effectiveness of the overall pipeline. Secondly, their integration is usually based on several tuning parameters whose optimal values are non-trivial to determine. Moreover, a performance drop in one module might cause the overall system to fail.

The aforementioned challenges can be addressed by leveraging DRL techniques~\cite{devo2020deep, devo2020towards, chaplot2020object}. A vast literature is available on DRL-based VAT approaches for ground vehicle systems. ~\cite{luo2019end} proposes an end-to-end deep neural network architecture to train a DRL agent in simulated environments and takes advantage of domain randomization in order to favor generalization to real-world scenarios. \cite{zhong2019ad} develops an asymmetric dueling training procedure employing an adversarial target that stimulates the development of an effective policy. In~\cite{dionigi2022vat}, the assumption of having the target within the camera FoV at the beginning of the maneuver is removed, so that the agent is able to explore an unknown environment, find the target and track it. All these approaches feature a discrete action space and therefore they cannot explore the full performance envelope of the vehicle. In fact, the resulting maneuvers are non-smooth and prone to losing visual contact with the target. An end-to-end architecture that exploits continuous actions is presented in~\cite{devo2021enhancing}.

Compared to ground robots, the design of learning-based policies for MAVs is significantly more challenging. In \cite{hwangbo2017control}, a multi-layer perceptron is coupled with a low-level PID controller in order to stabilize the MAV hovering configuration. This method employs absolute position measurements provided by motion capture system, and does not address the VAT problem. A VAT solution is proposed in~\cite{sampedro2018image} to allow a MAV to fly and track a moving object.
In particular, the control system of the MAV is designed to track ground targets by processing down-looking images, which precludes the application of the method to scenarios featuring front-looking cameras and flying targets.
\cite{xi2021anti} presents an active tracking module for MAVs equipped with a pan-tilt camera that is able to track a person in various complex scenes. Nonetheless, the MAV dynamics are not fully exploited in the design of the control policy and the action space is discrete, which poses a hard limit on the achievable performance.
A continuous action space is considered in~\cite{zhao2021active}, where a RL-based policy is coupled with a low-level PID control layer. However, the positioning of the MAV is constrained to a plane and thus the tracker is not free to move in 3D. Very few studies addressed the VAT problem for MAVs without relying on restrictive assumptions on the motion of the target-tracker pair. The recent work \cite{wang2023image} tackles this problem by adopting an image-based visual servoing approach that features a modular design similar to those discussed at the beginning of this section. Nevertheless, such a design leads to position and orientation errors in the order of 1 m and 0.1 rad, respectively, and it requires full attitude information.

\subsection{Contribution} \label{contribs}
As highlighted by the previous literature review, an increasing number of studies is focusing on VAT in the context of MAV applications. Model-based techniques (see, \eg \cite{wang2023image}) present design and integration issues that inherently limit their performance and entail tracking errors that may limit their applicability. On the other hand, existing learning-based approaches are affected by different constraints: (i) the target lies on a plane~\cite{sampedro2018image}; (ii) the tracker is controlled by discrete actions~\cite{xi2021anti}; (iii) the agent is trained with continuous actions that are confined to a subset of the tracker action space~\cite{zhao2021active}. To overcome these limitations, in this paper we provide the following contributions:
\begin{itemize}
    \item We propose D-VAT, a novel end-to-end DRL continuous control model for VAT applications involving MAVs.
    \item The proposed DRL policy directly maps RGB image data into thrust and angular velocity commands, and does not make restrictive assumptions on the trajectories of both the tracker and the target.
    \item We show the benefits of D-VAT by comparing it against different model-based and data-driven SotA approaches. Our approach outperforms the baselines also in scenarios that differ substantially from the training ones, \changes{and can be deployed on a real platform, thus} demonstrating remarkable generalization capabilities. 
\end{itemize}

\section{Preliminary Definitions} \label{sec:preliminary}
The optimization of RL models requires a significant number of interactions with the environment and this number becomes massive when deep approximators come into play. In practice, this excludes the possibility of using real MAVs to collect interaction episodes, both for efficiency and safety reasons. 
To overcome this issue, highly photorealistic simulation frameworks can be used to generate an unlimited amount of episodes and train the DRL models without any physical risk to the vehicle. In this work, we follow this practice and optimize our D-VAT model in simulated environments.
Before detailing the characteristics of D-VAT and its training procedure, in this section we describe the dynamic model which is integrated into the simulation engine to generate realistic motions. In particular, we follow~\cite{mueller2015computationally} and consider a surrogate model in which the tracker is controlled by thrust and angular velocity inputs. The model is given by:
\begin{equation}\label{sysmodel}
\begin{array}{c c l}
\ddot{p}&=& \dfrac{f}{m} R_3 -g\\[2mm]
\dot{R}&=&R\,[\omega]_\times 
\end{array} 
\end{equation}
In system \eqref{sysmodel}, $p$ and $R$ are the tracker absolute position and orientation, while $m$ and $g=[0\;0\;9.8]^\top \SI{}{\meter\per\second\squared}$ are the vehicle mass and the gravity vector, respectively. Moreover, $f$ and $\omega$ indicate the collective thrust and the angular velocity inputs. The notation $[\omega]_\times$ refers to the skew-symmetric representation of vector $\omega=[\omega_x\,\omega_y\,\omega_z]^T$. Since our DRL optimization framework is discrete-time, we apply a zero-order-hold discretization to system \eqref{sysmodel} and denote by $z(k)$ the value taken by a signal $z(t)$ at the sampling instant $t=k t_s$, where $t_s$ the sampling time. The motion of the target is modeled by a parameterized class of trajectories denoted by $p_r(k)$, as detailed in Section \ref{sec:env}. 
It is important to highlight that D-VAT is trained in a model-free manner and has no explicit information about the dynamics \eqref{sysmodel}. The simulation model is only used to generate realistic MAV trajectories. 

\section{Approach}\label{sec:approach}

\subsection{Problem Formulation}
\label{sec:problem_formulation}

The goal of VAT is to control the motion of a tracker agent equipped with a vision sensor, so as to maintain the target within the FoV of the camera and at a predefined distance.
In this paper, we assume that both the tracker and the target are MAVs that are free to move in 3D. 
The vision sensor is an RGB camera whose reference frame is coincident with the tracker body-fixed frame. In particular, the optical axis is aligned with the $x$-axis direction. At the beginning of the VAT task, the target is located ahead of the tracker (within the camera FoV), and starts moving along a time-varying trajectory. The tracker employs only the image stream coming from its front camera as a source of information and computes the thrust and angular velocity commands needed to meet the control goal. Similarly to other complex navigation and control tasks, VAT can be tackled by formulating a suitable reinforcement learning (RL) problem \cite{sutton2018reinforcement}. In particular, we treat the tracker as an RL agent which repeatedly interacts with an environment over a series of independent episodes. For each discrete timestep, the agent receives an observation $o(k)$, a reward $r(k)$, and produces an action $u(k)$. The observation is given by the aforementioned sequence of camera images, while the action is a continuous command that specifies the thrust and the angular velocity of the tracker MAV, \ie $u(k)=(f(k),\omega(k))$. The reward is defined in Section \ref{sec:optimization}). 

\subsection{Deep Reinforcement Learning Strategy} \label{sec:drl_strategy}

\begin{figure*}[t]
    \centering
    \includegraphics[width=\textwidth]{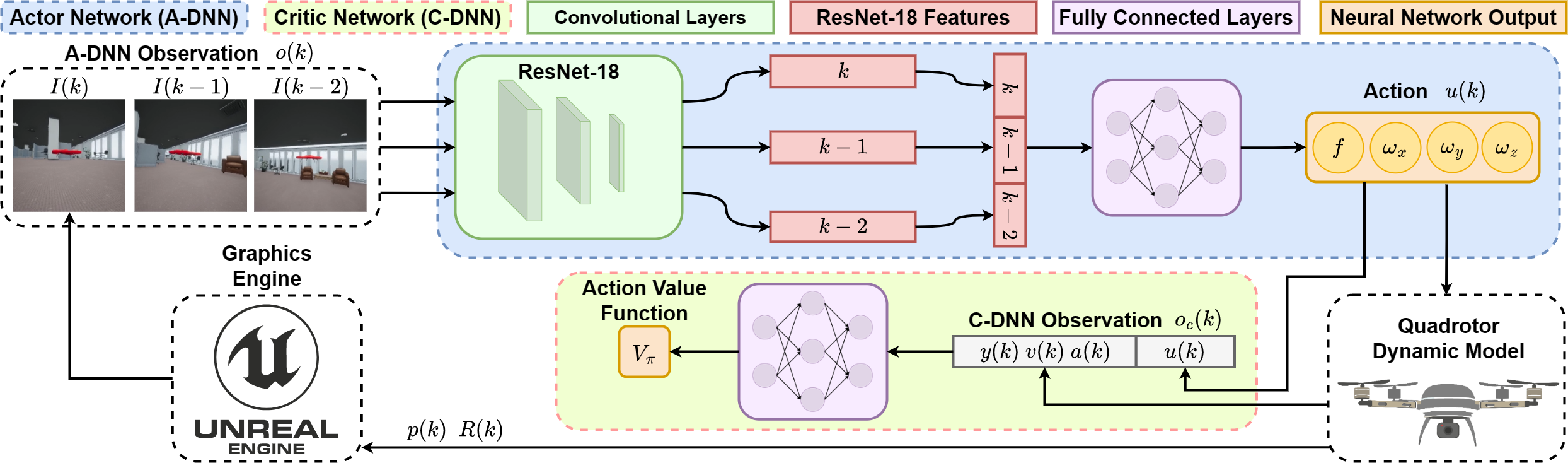}
    \caption{Overview of the proposed D-VAT architecture. The A-DNN (highlighted in blue) processes a batch of collected RGB images and computes the body-rate and thrust commands fed to the {tracker} MAV. The state of the tracker is updated according to the dynamic model \eqref{sysmodel} and the resulting pose is employed by the graphics engine to render the next image. The C-DNN (colored in light green) is instead provided with privileged information (relative position, velocity and acceleration) to facilitate the estimation of the action value function during training.}
    \vspace{-1em}
    \label{fig:arch} 
\end{figure*}

 The proposed end-to-end VAT strategy relies on a monocular setup and requires only an RGB image stream collected by the onboard camera to directly compute the MAV control commands. RGB images are partial observations of the full MAV state and are composed  of a large number of pixels that form a huge observation space. For this reason, it is not viable to train the agent using classical RL algorithms, and more advanced solutions based on Deep Neural Network (DNN) approximators must be applied. In particular, we adopt the \textit{asymmetric actor-critic} formulation \cite{pinto2017asymmetric, dionigi2022vat}. According to this framework \cite{sutton2018reinforcement}, we design two different DNN architectures for the \textit{actor} (A-DNN) and for the \textit{critic} (C-DNN). The former learns the optimal policy $u(k)=\pi(o(k))$ with respect to the given task, while the latter aims to evaluate such a policy during the training phase.
The asymmetric structure of this framework allows the critic network to be fed with more privileged information than the actor network, thus stimulating the development of an effective policy evaluation.
It is worth remarking that the 
A-DNN 
is the only agent operating at inference time.

The A-DNN is a convolutional neural network composed of a ResNet18 \cite{he2016deep} and three additional hidden layers, each one characterized by 512 neurons and ReLU activations.
In order to learn temporal relations, the proposed A-DNN design processes a sequence of $H$ front-view camera images. This turned out to play a key role in improving the tracking performance. The image sequence is given by 
\begin{equation}
\label{eq:observation_actor}
o(k)\!=\!\left[
\begin{array}{c c c c}
I(k)&
I(k - 1)&
\dots &
I(k-H+1)
\end{array}\right]^T,
\end{equation}
where $I(k)$ is the RGB frame acquired at the $k$-th time step. 
Moreover, the A-DNN extracts 512 visual features from each image through its convolutional block. Subsequently, the $H \times 512$ features are concatenated and fed to the linear layers to compute the action. 
The control actions are saturated to be consistent with the physical characteristics of MAV actuators. In particular, a $\tanh$ saturation is adopted to confine the action values computed by the A-DNN within prescribed limits (see angular rate and thrust limits in Table \ref{table:hyper}).

The C-DNN design consists of a fully connected neural network with three hidden layers, each one composed of 256 neurons and ReLU activations. The correct selection of the inputs to the  C-DNN  is, in general, nontrivial. In this work, we explored different possibilities and selected the input set that we found to be the most informative without unnecessarily increasing the network complexity. In particular, we define the observation of the C-DNN as a vector $o_c(k)$ representing the relative state as follows:
\begin{equation}
\label{eq:observation_critic}
o_c(k)\!=\!\left[
\begin{array}{c}
y(k)\\
v(k)\\
a(k)
\end{array}\right]
\!=\!\left[
\begin{array}{c}
R(k)^T [p_r(k)-p(k)]\\
R(k)^T [\dot{p}_r(k)- \dot{p}(k)]\\
R(k)^T [\Ddot{p}_r(k)- \Ddot{p}(k)]
\end{array}\right],
\end{equation}
where $y(k)$, $v(k)$ and $a(k)$ denote respectively the position, velocity and acceleration of the target relative to the tracker, expressed in the tracker body-fixed frame. The C-DNN 
output is a scalar representing the estimated \textit{action-value} $Q_\mathbf{\pi}(o_c(k), u(k))$. The overall design is illustrated in Fig.~\ref{fig:arch}.

\subsection{Optimization} 
\label{sec:optimization}
The A-DNN and the C-DNN are both trained by using the popular RL-based Soft Actor-Critic (SAC) framework \cite{haarnoja2018soft}, where the reward signal $r(k)$ is specifically designed to address the VAT problem in MAVs scenarios, taking into account the distinctive characteristics and requirements of the considered control task. In particular, the main control objective is to align the target with the center of the tracker camera FoV while keeping a predefined distance between the two vehicles. To this purpose, the reward is defined as: 
\begin{equation} \label{rew:track}
    r_{e}(k) = (r_x(k) \, r_y(k) \, r_z(k))^{\beta},
\end{equation}
where $\beta>0$ is a suitable exponent and 
\begin{equation} \label{rew:axis}
\begin{array}{l l l}
r_x &=& \max(0, 1 - \left | y_{x}(k) - d_r \right |), \\[2mm] 
r_y &=& \max\left(0, 1 - \left | \frac{2}{A_\text{FoV}}\arctan\left(\frac{y_y(k)}{y_x(k)}\right) \right |\right), \\[2mm]
r_z &=& \max\left(0, 1 - \left | \frac{2}{A_\text{FoV}}\arctan\left(\frac{y_z(k)}{y_x(k)}\right) \right |\right).
\end{array}
\end{equation}
In Eq. \eqref{rew:axis}, $r_x$ is maximal when the first entry of $y(k)=[y_x(k)\,y_y(k)\,y_z(k)]^T$ matches the desired distance $d_r$ to the target ($d_r$ is specified along the $x$-axis of the body-fixed frame, which is assumed coincident with the optical axis). Moreover, $r_y$ and $r_z$ are functions that encourage the agent to keep the target at the center of the image plane and thus away from the camera FoV limits, being $A_{\text{FoV}}$ the FoV amplitude in radians. The reward term $r_{e}(k)$ in \eqref{rew:track} is clipped in the interval $\left [ 0,\; 1\right ]$ to favor the learning process, and it is maximal ($r_e=1$) when the VAT goal is achieved.

Two additional reward terms are included in the formulation to optimize also the control effort and the MAV linear velocity. In particular, we define a velocity penalty $r_{v}$ and a control effort penalty $r_{u}$ as follows:
\begin{equation} \label{pen:vel}
    r_{v}(k) = \dfrac{\| v(k) \|}{1 + \| v(k) \|}, \;\; r_{u}(k) = \dfrac{\| u(k) \|}{1 + \| u(k) \|}.
\end{equation}
Collision avoidance constraints are taken into consideration by penalizing the RL agent whenever $\| y(k) \| < d_{m}$, where $d_{m}$ is the minimum distance allowed.

The reward function is obtained by adding up all the above contributions, which results in:
\begin{equation} \label{rew:total}
r(k) = \left\{\begin{array}{l c}
\!\!r_{e}(k) \!-\! k_v  r_{v}(k)\! - \!k_u r_{u}(k) & \| y(k) \| > d_{m} \\[1mm] 
\!\!-k_c & \text{otherwise},
\end{array}\right. 
\end{equation}
where $k_c$ is a large positive constant and $k_v>0$, $k_u>0$ are weighting parameters that we carefully balance in order to realize a high-performance control policy.

\subsection{Training Environment} 
\label{sec:env}

\begin{figure}[t]
\centering

\includegraphics[width=.29\columnwidth]{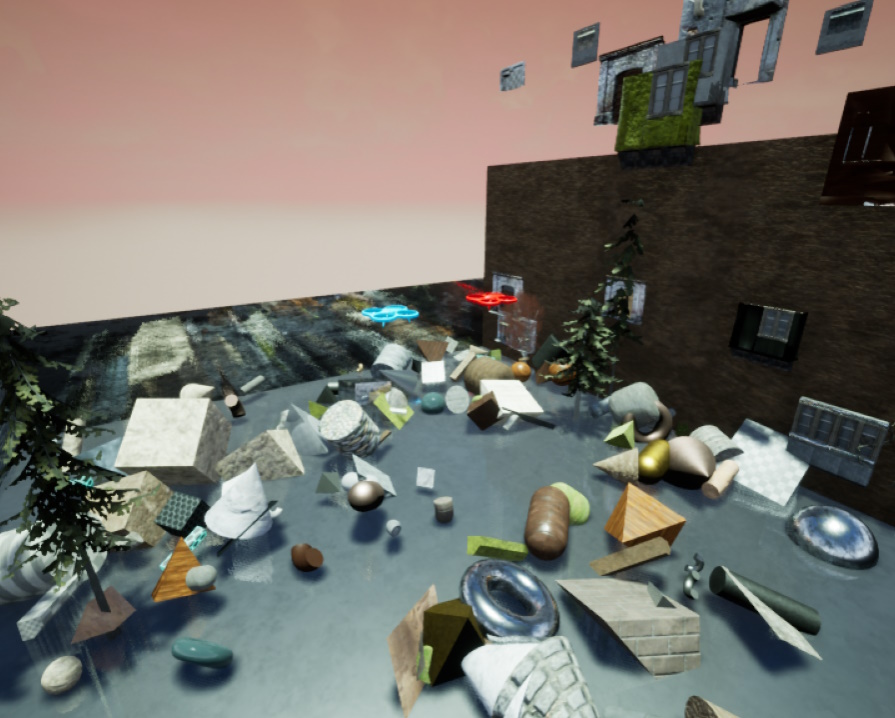}
\includegraphics[width=.29\columnwidth]{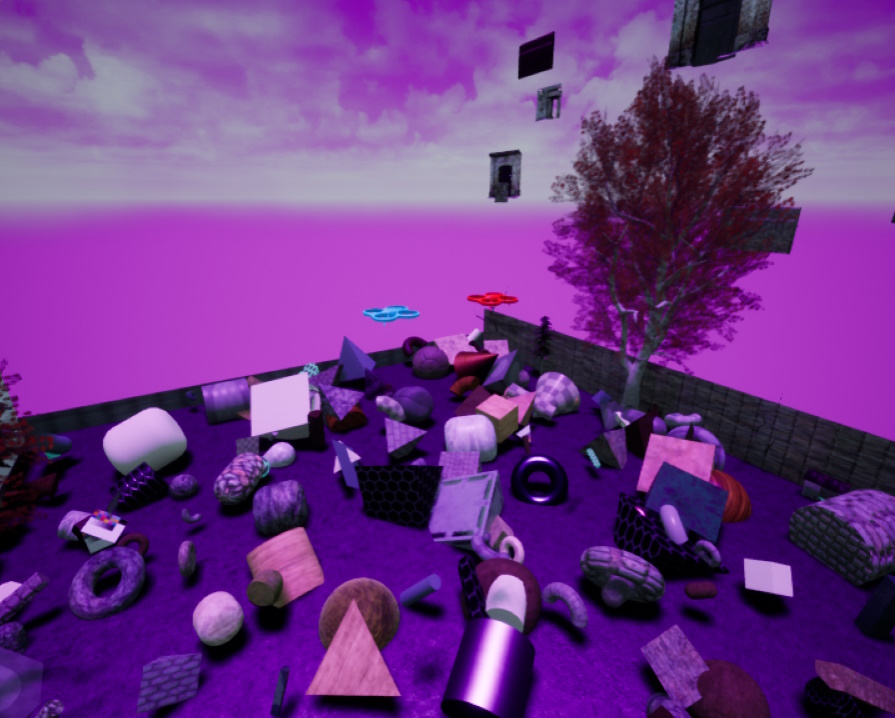}
\includegraphics[width=.29\columnwidth]{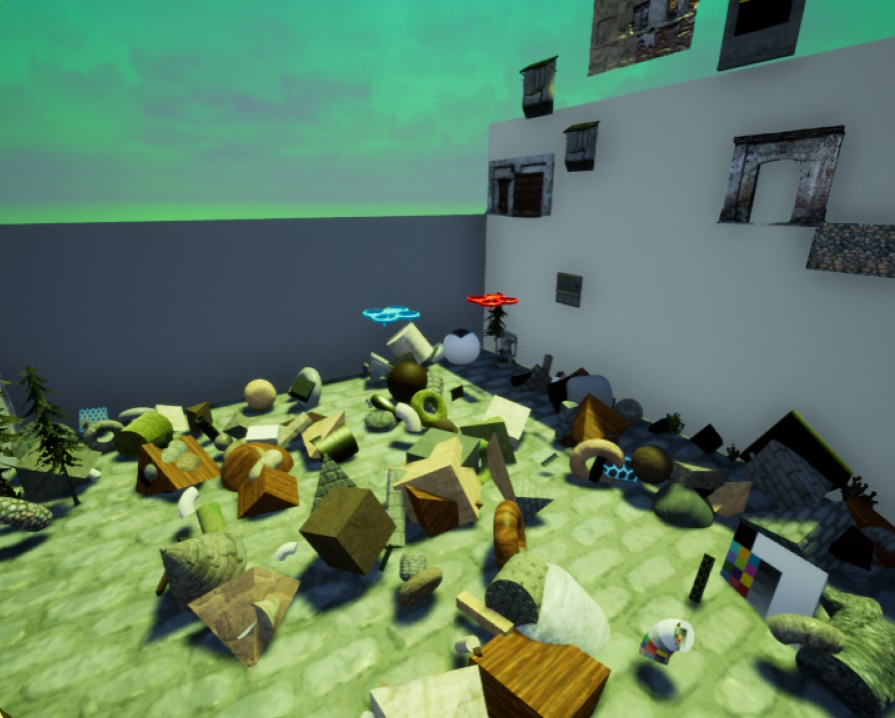}
\vspace{-1.0em}
\caption{\changes{Examples of the training environment randomization.  The tracker (blue) and the target (red) MAVs are spawned in a large room with randomized characteristics.}}
\vspace{-1.0em}
\label{fig:trainEnvs}

\end{figure}

To build a simulated environment suitable for training the DRL agent, we employ Unreal Engine 4 (UE)\footnote{https://www.unrealengine.com}, a popular graphics engine that provides advanced photorealism capabilities.
Background patterns and objects in the simulated scene play an important role since the agent's first aim is to learn how to effectively and reliably distinguish the target from the scenery. To this aim, we choose a large furnished room (see Fig. \ref{fig:trainEnvs}) as the environment used for optimization. Moreover, we exploit
\textit{domain randomization} \cite{tobin2017domain} to facilitate the development of the robustness and generalization capabilities needed to directly deploy the DRL agent in more complex photorealistic environments. Before the beginning of each training episode, the room characteristics are randomly altered in light conditions, furniture, and texture patterns, including those of walls and floors. As soon as the room is randomized, the {target} is spawned in front of the {tracker}, whose initial position is randomly set. The target then starts moving along a sinusoidal trajectory parameterized as follows: 
\begin{equation*}\label{referencetrajectory}
p_r(k) = p_r(0)+
\begin{bmatrix}
    A_x  \sin(2\pi f_x  k + \phi_x)\\ 
    A_y  \sin(2\pi f_y  k + \phi_y)\\ 
    A_z  \sin(2\pi f_z  k + \phi_z)    
\end{bmatrix}
-
\begin{bmatrix}
    A_x  \sin(\phi_x)\\ 
    A_y  \sin(\phi_y)\\ 
    A_z  \sin(\phi_z)    
\end{bmatrix},
\end{equation*}
where ($A_x$, $A_y$, $A_z$), ($f_x$, $f_y$, $f_z$), ($\phi_x$, $\phi_y$, $\phi_z$) are respectively the amplitude, frequency, and phase, which are uniformly sampled in the intervals specified in Table \ref{table:hyper} to produce a different trajectory for each episode.

We exploit parallel training to accelerate the optimization. To this purpose, we instantiate multiple rooms as described above, each with dedicated target-tracker pair and an independent randomization sequence.  The environments used to test our approach are described in Section \ref{sec:exp_setup}.

\section{Experiments} \label{sec:experiments}
In this section, we detail the implementation of our approach and discuss the experimental campaign.

\subsection{Experimental Setup}
\label{sec:exp_setup}

The A-DNN and C-DNN have been optimized by using the Stable-Baselines3 \cite{stable-baselines3} implementation of SAC, which we customize
to extend it to the \textit{asymmetric actor-critic} formulation of our approach. 
The networks have been optimized for approximately 18,000 episodes executed in 6 parallel environments, using the Adam optimizer with a learning rate of 0.0003, a discount factor $\gamma$ of 0.99, and a batch size of 64. Each training episode has a maximum duration of $40$ s, and the observation sequence length for the A-DNN is set to $H=3$. The other hyper-parameters and settings are reported in Table \ref{table:hyper}. The training process is performed on a workstation equipped with 2 × NVIDIA RTX 2080Ti with 11GB of VRAM, an Intel Core processor i7-9800X (3.80GHz ×16) and 64 GB of DDR4 RAM. 

\begin{figure}[t]
\centering

\includegraphics[width=.29\columnwidth]{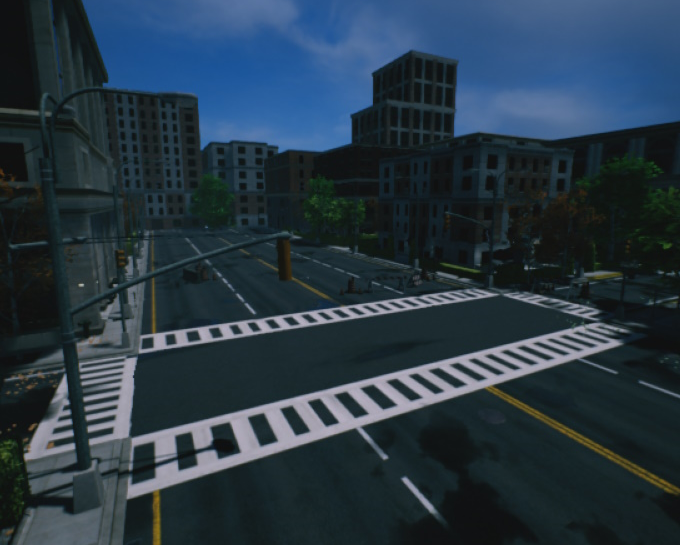}
\includegraphics[width=.29\columnwidth]{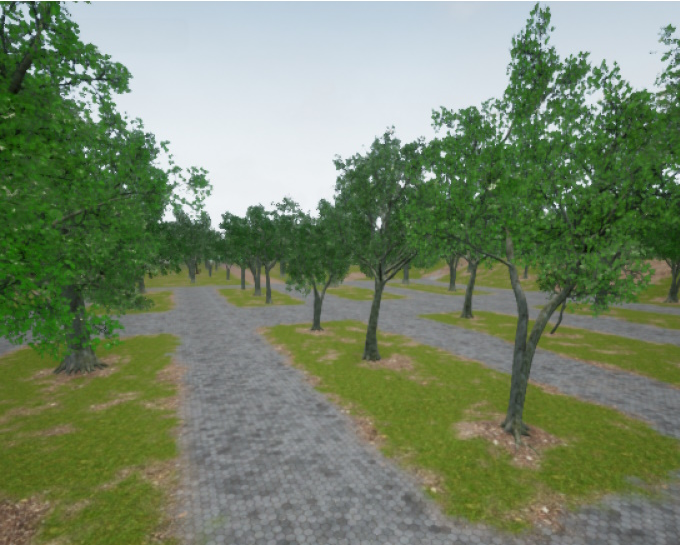}
\includegraphics[width=.29\columnwidth]{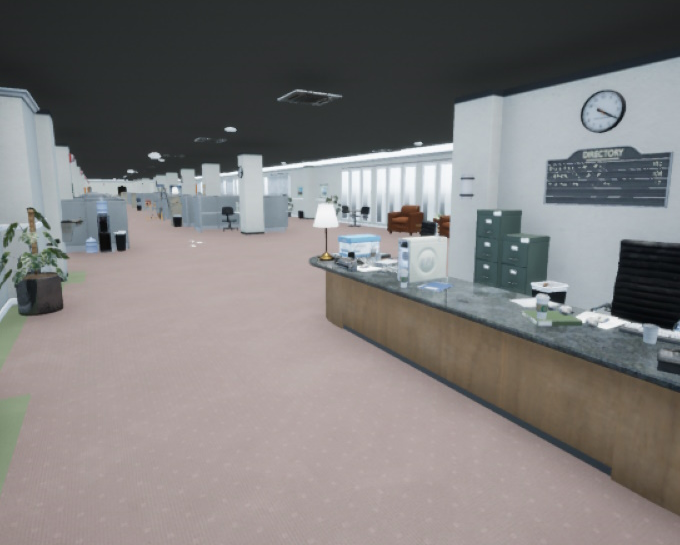}
\vspace{-1.0em}
\caption{\changes{Images from the photo-realistic environments employed to test the generalization capabilities of D-VAT. From left to right: an urban setting (Urban), a park environment (Park), and an office space (Office).}} 
\vspace{-1.5em}
\label{fig:realEnvs}

\end{figure}
\begin{table}[t]
\renewcommand{\arraystretch}{1.3}
\caption{Hyperparameters and Settings\vspace{-1.0em}}
\begin{center}
\resizebox{0.62\columnwidth}{!}{
\label{table:hyper}
\centering
\begin{tabular}{cc}
\hline
\textbf{Hyperparameter} & \textbf{Value} \\
\hline
Tracker MAV mass $m$ & $1$ kg\\
Camera FoV amplitude $A_\text{FoV}$ & $\pi/2$ rad\\
Reward exponent $\beta$ & $1/3$ \\
Reward coefficients $k_v, k_u$ & $0.4$ \\
Reward constant $k_c$ & $10$ \\
Distance setpoint $d_{r}$ & $ 0.5$ m \\
Min. allowed distance $d_{m}$ & $0.3$ m \\
Target traj. amplitude $A_x, A_y, A_z $ & $\left[1, 2.5\right]$ m \\
Target traj. frequency $f_x, f_y, f_z$ & $\left[0.04, 0.25\right]$ Hz \\
Target traj. phase $\phi_x, \phi_y, \phi_z$ & $\left[0, 2\pi\right]$ \\
Angular rate limits on $\omega$ & $\left[-4, 4\right]$ rad/s \\
Thrust limits on $f$ & $\left[0.1, 20.1\right]$ N \\
\hline
\end{tabular}
}
\end{center}
\vspace{-2.8em}
\end{table}

Our approach is tested on two environment classes: the first one contains scenes similar to those used during the training phase, although with different room shapes, objects disposition, and textures (we refer to these scenes as Box Environments). The second is, instead, aimed at testing the generalization capabilities of D-VAT and has more complex and photo-realistic environments, \ie an outdoor urban scenario (Urban), an outdoor park environment (Park), and an indoor scene of an office building (Office). These are depicted in Fig.~\ref{fig:realEnvs} and are significantly different from the ones used to train our model. 

We run a total of 20 maneuver realizations for each test environment. In each run, the tracker is spawned at a random initial position, while the target is initially placed in front of the tracker at the optimal distance. To assess the generalization capabilities of our approach, we test also target trajectories that differ from the training ones. In particular, we consider constant setpoints and rectilinear trajectories with different shapes such as ramp-like and cubic. In the following, the D-VAT agent is compared to the SotA baselines described hereafter. 

\begin{table*}[t]
\renewcommand{\arraystretch}{1.3}
\centering
\caption{Experimental results comparing our approach against the baseline in the box environments (similar to those used during training) and in the photorealistic scenarios (Urban, Park and Office)\vspace{-1.0em}}
\resizebox{1.6\columnwidth}{!}{
\label{tab:results}
\begin{tabular}{|c|c|c|c|c|c|c|c|c|c|c|c|c|c|c|c|c|}
\hline
\multirow{3}{*}{Method} & \multicolumn{16}{c|}{Experimental Scenarios and Metrics}\\
\cline{2-17}
& \multicolumn{4}{c|}{Box Environments} & \multicolumn{4}{c|}{Urban} & \multicolumn{4}{c|}{Park} & \multicolumn{4}{c|}{Office}\\
\cline{2-17}
& $P_{\theta}$ & $P_{\varphi}$ & $P_{\rho}$ & $P_{c}$ & $P_{\theta}$ & $P_{\varphi}$ & $P_{\rho}$ & $P_{c}$ & $P_{\theta}$ & $P_{\varphi}$ & $P_{\rho}$ & $P_{c}$ & $P_{\theta}$ & $P_{\varphi}$ & $P_{\rho}$ & $P_{c}$ \\
\hline
AOT \cite{luo2019end} & $ \cellcolor{red!30} 0.03$ & $ \cellcolor{red!30} 0.04$ & $ \cellcolor{red!30} 0.03$ & $ \cellcolor{red!30} 0.03$ & $ \cellcolor{red!30} 0.03$ & $ \cellcolor{red!30} 0.04$ & $ \cellcolor{red!30} 0.03$ & $ \cellcolor{red!30} 0.03$ & $ \cellcolor{red!30} 0.03$ & $ \cellcolor{red!30} 0.04$ & $ \cellcolor{red!30} 0.03$ & $ \cellcolor{red!30} 0.03$ & $ \cellcolor{red!30} 0.03$ & $ \cellcolor{red!30} 0.04$ & $ \cellcolor{red!30} 0.03$ & $ \cellcolor{red!30} 0.03$ \\
\hline
AD-VAT+ \cite{zhong2019ad} & $ \cellcolor{red!30} 0.03$ & $ \cellcolor{red!30} 0.03$ & $ \cellcolor{red!30} 0.03$ & $ \cellcolor{red!30} 0.03$ & $ \cellcolor{red!30} 0.03$ & $ \cellcolor{red!30} 0.03$ & $ \cellcolor{red!30} 0.03$ & $ \cellcolor{red!30} 0.03$ & $ \cellcolor{red!30} 0.03$ & $ \cellcolor{red!30} 0.03$ & $ \cellcolor{red!30} 0.03$ & $ \cellcolor{red!30} 0.03$ & $ \cellcolor{red!30} 0.03$ & $ \cellcolor{red!30} 0.03$ & $ \cellcolor{red!30} 0.03$ & $ \cellcolor{red!30} 0.03$ \\
\hline
C-VAT \cite{devo2021enhancing} & $ \cellcolor{red!30} 0.05$ & $ \cellcolor{red!30} 0.04$ & $ \cellcolor{red!30} 0.04$ & $ \cellcolor{red!30} 0.04$ & $ \cellcolor{red!30} 0.05$ & $ \cellcolor{red!30} 0.04$ & $ \cellcolor{red!30} 0.04$ & $ \cellcolor{red!30} 0.04$ & $ \cellcolor{red!30} 0.05$ & $ \cellcolor{red!30} 0.04$  & $ \cellcolor{red!30} 0.04$ & $ \cellcolor{red!30} 0.04$ & $ \cellcolor{red!30} 0.05$ & $ \cellcolor{red!30} 0.04$ & $ \cellcolor{red!30} 0.04$ & $ \cellcolor{red!30} 0.04$\\
\hline
SiamRPN++ \cite{li2019siamrpn++} LQG & $ \cellcolor{yellow!40} 0.27$ & $ \cellcolor{yellow!40} 0.20$ & $ \cellcolor{yellow!40} 0.20$ & $ \cellcolor{yellow!40} 0.22$ & $ \cellcolor{yellow!40} 0.10$ & $ \cellcolor{yellow!40} 0.08$ & $ \cellcolor{yellow!40} 0.07$ & $ \cellcolor{yellow!40} 0.09$ & $ \cellcolor{yellow!40} 0.37$ & $ \cellcolor{yellow!40} 0.27$  & $ \cellcolor{yellow!40} 0.27$ & $ \cellcolor{yellow!40} 0.30$ & $ \cellcolor{yellow!40} 0.11$ & $ \cellcolor{yellow!40} 0.08$ & $ \cellcolor{yellow!40} 0.09$ & $ \cellcolor{yellow!40} 0.09$\\
\hline
SiamRPN++ \cite{li2019siamrpn++} PID & $ \cellcolor{yellow!40} 0.65$ & $ \cellcolor{yellow!40} 0.65$ & $ \cellcolor{yellow!40} 0.55$ & $ \cellcolor{yellow!40} 0.62$ & $ \cellcolor{yellow!40} 0.57$ & $ \cellcolor{yellow!40} 0.56$ & $ \cellcolor{yellow!40} 0.46$ & $ \cellcolor{yellow!40} 0.53$ & $ \cellcolor{yellow!40} 0.94$ & $ \cellcolor{yellow!40} 0.93$  & $ \cellcolor{yellow!40} 0.76$ & $ \cellcolor{yellow!40} 0.88$ & $ \cellcolor{yellow!40} 0.78$ & $ \cellcolor{yellow!40} 0.75$ & $ \cellcolor{yellow!40} 0.67$ & $ \cellcolor{yellow!40} 0.73$\\
\hline
D-VAT (Our) & $ \cellcolor{green!40} 0.94$ & $ \cellcolor{green!40} 0.94$ & $ \cellcolor{green!40} 0.86$ & $ \cellcolor{green!40} 0.91$ & $ \cellcolor{green!40} 0.94$ & $ \cellcolor{green!40} 0.94$ & $ \cellcolor{green!40} 0.91$ & $ \cellcolor{green!40} 0.93$ & $ \cellcolor{green!40} 0.95$ & $ \cellcolor{green!40} 0.95$  & $ \cellcolor{green!40} 0.92$ & $ \cellcolor{green!40} 0.94$ & $ \cellcolor{green!40} 0.95$ & $ \cellcolor{green!40} 0.95$ & $ \cellcolor{green!40} 0.86$ & $ \cellcolor{green!40} 0.92$\\
\hline
\end{tabular}}
\vspace{-1.em}
\end{table*}
\begin{table*}[t]
\renewcommand{\arraystretch}{1.1}
\centering
\caption{ Experimental results obtained for different peak velocities of the target \vspace{-1.0em}}
\resizebox{1.3\columnwidth}{!}{
\label{tab:velocity}
\begin{tabular}{|c|c|c|c|c|c|c|c|c|c|c|c|c|}
\hline
\multirow{3}{*}{Method} & \multicolumn{12}{c|}{Peak Velocity of the Target}\\
\cline{2-13}
& \multicolumn{4}{c|}{$0.5$ m/s} & \multicolumn{4}{c|}{$1$ m/s} & \multicolumn{4}{c|}{$2$ m/s}\\
\cline{2-13}
& $P_{\theta}$ & $P_{\varphi}$ & $P_{\rho}$ & $P_{c}$ & $P_{\theta}$ & $P_{\varphi}$ & $P_{\rho}$ & $P_{c}$ & $P_{\theta}$ & $P_{\varphi}$ & $P_{\rho}$ & $P_{c}$\\
\hline
SiamRPN++ \cite{li2019siamrpn++} LQG & $ \cellcolor{green!40} 0.98$ & $ \cellcolor{red!30} 0.87$ & $ \cellcolor{red!30} 0.88$ & $ \cellcolor{red!30} 0.91$ & $ \cellcolor{red!30} 0.47$ & $ \cellcolor{red!30} 0.33$ & $ \cellcolor{red!30} 0.35$ & $ \cellcolor{red!30} 0.38$ & $ \cellcolor{red!30} 0.04$ & $ \cellcolor{red!30} 0.03$  & $ \cellcolor{red!30} 0.02$ & $ \cellcolor{red!30} 0.03$\\
\hline
SiamRPN++ \cite{li2019siamrpn++} PID &$ \cellcolor{yellow!40} 0.97$
& $ \cellcolor{green!40} 0.96$ & $ \cellcolor{green!40} 0.95$ & $ \cellcolor{green!40} 0.96$ & $ \cellcolor{yellow!40} 0.94$ & $ \cellcolor{yellow!40} 0.93$ & $ \cellcolor{yellow!40} 0.78$ & $ \cellcolor{yellow!40} 0.88$ & $ \cellcolor{yellow!40} 0.09$ & $ \cellcolor{yellow!40} 0.09$  & $ \cellcolor{yellow!40} 0.06$ & $ \cellcolor{yellow!40} 0.08$\\
\hline
D-VAT (Our) & $ \cellcolor{red!30} 0.96$ & $ \cellcolor{green!40} 0.96$ & $ \cellcolor{yellow!40} 0.90$ & $ \cellcolor{yellow!40} 0.94$ & $ \cellcolor{green!40} 0.95$ & $ \cellcolor{green!40} 0.95$ & $ \cellcolor{green!40} 0.90$ & $ \cellcolor{green!40} 0.93$ & $ \cellcolor{green!40} 0.91$ & $ \cellcolor{green!40} 0.89$  & $ \cellcolor{green!40} 0.84$ & $ \cellcolor{green!40} 0.88$\\
\hline
\end{tabular}}
 \vspace{-1.8em}
\end{table*}

\subsection{Baselines}
\label{sec:baselines}

 \textbf{Active Object Tracking (AOT)} \cite{luo2019end}. In this approach, the agent is trained to track predefined target trajectories by using discrete actions. To comply with the dynamic model \eqref{sysmodel}, which takes as input the collective thrust and angular velocity of the MAV, we define the action set as follows: \{$+\Delta\omega_x$, $\!-\Delta\omega_x$, $+\Delta\omega_y$, $\!-\Delta\omega_y$, $+\Delta\omega_z$, $\!-\Delta\omega_z$, $+\Delta f$, $\!-\Delta f$, $no\_op$ \}, where the operator $\Delta$ indicates a fixed increment of thrust or angular velocity and $no\_op$ prescribes a zero thrust or angular velocity increment. The size of the $\Delta$ increments has been manually tuned to meet the task specifications. 

\textbf{AD-VAT+} \cite{zhong2019ad}. The model policy is learned during the adversarial dueling against the target, which is itself an RL agent. This approach employs the same discrete action space as the AOT baseline.

\textbf{C-VAT} \cite{devo2021enhancing}. The model is optimized using a target that is randomly spawned in the surrounding of the tracker. In particular, a heuristic trajectory generator (HTG) is combined with a suitable set of auxiliary losses in order to facilitate the convergence of the training process. Herein, we implement the HTG with a Linear Quadratic Gaussian (LQG) controller that exploits ground truth pose information to control the tracker so as to achieve the VAT goal. Moreover, the auxiliary losses in \cite{devo2021enhancing} have been extended to a 3D environment.

\textbf{SiamRPN++ PID}. This modular baseline combines the object tracker SiamRPN++  \cite{li2019siamrpn++} with a standard MAV control architecture featuring two Proportional-Integral-Derivative (PID) feedback loops. 
In order to achieve the VAT goal, the outer loop processes the bounding box information provided by SiamRPN++ (\ie position and size of the bounding box enclosing the target) to compute roll, pitch, yaw, and thrust signals that are are fed to the inner (attitude control) loop. 
The PID parameters have been tuned using a trial and error approach on relevant scenarios, so as to achieve a suitable trade-off between reactivity to tracking errors and sensitivity to noise. The inner loop needs attitude information and, in our tests, we provide the ground-truth attitude angles returned by the simulator. This baseline is favored with respect to D-VAT because it has access to privileged information, \ie the attitude of the MAV.


\textbf{SiamRPN++ LQG}. This modular baseline combines SiamRPN++ with a model-based design that couples feedback linearization and a linear control law (see, \eg \cite{leomanni23robust}). In particular, we adopt a Linear-Quadratic-Gaussian (LQG) design. The resulting policy uses the bounding box information to regulate directly the thrust and angular velocity of the tracker so as to meet the VAT objective. The LQG weights have been tuned extensively to achieve a fair trade-off between performance and robustness. This baseline requires attitude information (to linearize the MAV dynamics by feedback) and hence it is favored with respect to D-VAT. 

\subsection{Metrics}
\label{sec:metrics}

To evaluate the performance of D-VAT against that of the baselines, we adapted the tracking metrics in \cite{devo2021enhancing, dionigi2022vat} to a 3D environment. For convenience, the metrics are defined by expressing the ground-truth position of the target relative to the tracker in a  spherical coordinate system, whose axes are aligned with those of the tracker body-fixed frame. The spherical coordinates are denoted by $(\rho, \theta, \varphi)$. The considered metrics are detailed below.

\textbf{Distance Score}: measures the ability of the tracker to maintain the desired distance from the {target}, as follows
\begin{equation*}
\small
	\tilde{P}_{\rho}(k) = 
	\begin{cases}
		\max\left(0, 1 - 2|\rho(k) - d_r|\right), & \mbox{if } \begin{array}{ll}
		     |\theta(k)| < \frac{A_\text{FoV}}{2} \\
		     |\varphi(k)| < \frac{A_\text{FoV}}{2}
		\end{array}\\
		0 & \mbox{otherwise}
	\end{cases}
\end{equation*}

\textbf{Elevation Score}: measures the ability of the tracker to maintain the {target} {vertically} aligned to the center of the FoV, as follows
\begin{equation*}
\small
	\tilde{P}_{\theta}(k) = 
	\begin{cases}
		\max\left(0, 1 - \frac{2|\theta(k)|}{A_\text{FoV}}\right), & \mbox{if } \begin{array}{ll}
		     |\varphi(k)| < \frac{A_\text{FoV}}{2}  \\
		     |\rho(k) - d_r| < 0.5 
		\end{array}\\
		0 & \mbox{otherwise}
	\end{cases}
\end{equation*}

\textbf{Azimuth Score}: measures the ability of the tracker to maintain the target {horizontally} aligned to the center of the FoV, as follows
\begin{equation*}
\small
	\tilde{P}_{\varphi}(k) = 
	\begin{cases}
		\max\left(0, 1 - \frac{2|\varphi(k)|}{A_\text{FoV}}\right), & \mbox{if } \begin{array}{l}
                         |\theta(k)| < \frac{A_\text{FoV}}{2}\\
                        |\rho(k) - d_r| < 0.5 
                        \end{array}\\
		0 & \mbox{otherwise}
	\end{cases}
\end{equation*}

\textbf{Total Score}: it is the arithmetic mean of the above metrics, given by $\tilde{P}_{c}(k) = (\tilde{P}_{\rho}(k) + \tilde{P}_{\theta}(k) + \tilde{P}_{\varphi}(k))/{3}$.

Notice that if $\tilde{P}_{\rho}(k)=1$, then the tracker is at the desired distance from the target. Moreover, if $\tilde{P}_{\theta}$ and $\tilde{P}_{\phi}$ are both equal to $1$, then the target centroid is at the center of the FoV. Summarizing, $\tilde{P}_{c}(k)=1$ when perfect visual tracking is achieved at step $k$. 

The metrics are averaged with respect to the episode time and across the 20 runs performed in each scenario, resulting in 
$
	P_{m} = \frac{1}{20 N_e} \sum_{i=1}^{20} \sum_{k=0}^{N_c-1} {}^{(i)\!}\tilde{P}_{m}(k) \;,
$
where $m \in \{\rho, \theta, \varphi,  c\}$, ${}^{(i)\!}\tilde{P}$ indicates that the performance is evaluated on the $i$-th run, and $N_c$ is the number of samples within the episode.

\begin{figure*}[t]
    \centering
    \includegraphics[width=\textwidth]{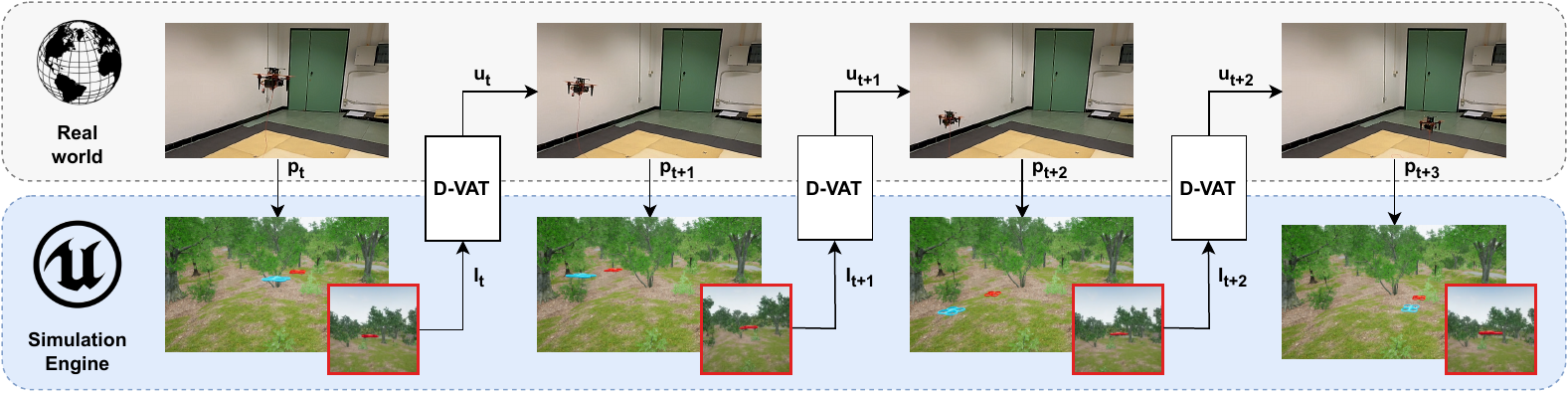}
    \caption{\changes{Mixed-Reality framework: the simulation engine renders the image and collects the RGB observation $I_t$ of the tracker MAV. D-VAT then predicts the control signal $u_t$ to follow the target drone. $u_t$ is directly employed to command the real drone.
    The new position of the real drone $p_{t+1}$ is used to update the position of the simulated tracker drone and collect the next RGB observation $I_{t+1}$.}}
    \vspace{-1em}
    \label{fig:mixed_reality} 
\end{figure*}

\subsection{Comparison Results} 
\label{sec:results}
The results of the experimental campaign are presented in Tables \ref{tab:results} and \ref{tab:velocity}. Our first important finding is that D-VAT outperforms all the baselines with respect to the performance metrics, and it is able to track the target by producing low-level control commands directly from RGB images. 
A visual inspection of the experiments (see the supplementary videos for qualitative results) shows that D-VAT is able to react promptly and effectively to the target movements. Specifically, it (i) computes fast maneuvers when the target approaches the boundary of camera FoV to avoid losing it, and (ii) provides a smooth control policy that is close to being optimal (\ie the target is almost always maintained at the center of the image plane and at the desired distance).

The learning-based approaches AOT, AD-VAT+ and C-VAT fail to converge to a suitable tracking policy. This could be explained by considering the high complexity of the task. AOT and AD-VAT+ are both strategies that rely on a discrete action space. Thus, they generate non-smooth control policies that struggle to maintain the target visibility
and might even result in unstable maneuvers that cause the target to disappear outside the FoV. Even C-VAT, despite being designed to provide continuous commands, fails to provide an efficient tracking policy. To explain this result, it is important to notice that the dimension of the MAV action space is doubled with respect to that of a planar ground robot (which is the platform considered in the original C-VAT work \cite{devo2021enhancing}). The increased complexity of the quadrotor dynamics 
make the model optimization more challenging and, in the case of C-VAT, this entails a large performance degradation.

The baselines that combine two separate modules, \ie an object detector and a controller (LQG or PID), are instead able to achieve better results. Nonetheless, the overall tracking performance is inferior to that of D-VAT. This can be attributed to the modular nature of these baselines. As the two components are designed independently, their coupling turns out to be inefficient and can cause the overall system to fail. In practice, this problem emerges since the controller, which has been designed under the assumption that the relative position is accurately known, is fed with position measurements extracted from the bounding box information provided by the object detector. These measurements, due to non-ideal image conditions or aggressive target maneuvers, might violate the design assumptions. This aspect becomes even more critical in realistic environments that are characterized by a high density of distracting objects in the background (\eg the photorealistic scenarios Urban and Office in Fig.~\ref{fig:realEnvs}). In this regard, it should be noted that the PID scheme, thanks to its more adaptable design, is more robust to model mismatch than the LQG counterpart.

On the other hand, thanks to the domain randomization strategy we employ, D-VAT has learned a tracking policy that can deal effectively with a wide range of scenarios and at the same time achieve high performance. This holds even when the visual conditions of the environment are very different from those employed in the training phase.
\changes{Moreover, to verify the visual robustness of D-VAT against dynamic objects, we employ a validation scenario featuring two moving items that occasionally appear in the tracker FoV. One of them shares the same shape as the target one but has a different color, while the other has a different shape but the same color as the target MAV. As shown in the attached video, the tracker agent is nearly unaffected by the presence of dynamic distracting objects, which proves that it did not overfit with respect to the object color or its shape individually. This is even more remarkable if we consider that no moving objects other than the target MAV are included in the training phase.}

To further study the comparison between D-VAT and the modular baselines, we run additional experiments by varying the maximum velocity of the target. We perform these experiments on a simplified scene with low amount of texture and no objects. In Table~\ref{tab:velocity}, it can be seen that for low target velocities, the modular baselines and D-VAT achieve similar performance. However, when the target performs faster and more aggressive trajectories, the performance of both the modular baselines decreases, while D-VAT tracking capabilities are almost unaffected. This suggests that the proposed learning-based approach is more robust and responsive in challenging scenarios where the ability of traditional control strategies may be limited.

\subsection{\changes{DRL Controller Validation with Mixed-Reality}}
\changes{To assess the sim-to-real adaptation capabilities of D-VAT we follow the strategy in \cite{devo2022autonomous} and design a Mixed Reality framework.
Specifically, we deploy the D-VAT model into a real MAV platform and employ UE for rendering purposes only. At each timestep the real tracking MAV and the simulator interact as follows: first, the current pose of the real MAV is provided to the simulator to update the position of its simulated counterpart. Then, the UE renders the RGB camera frames and provides them to D-VAT to compute the control commands. These are sent back to the real platform that executes them (see Figure \ref{fig:mixed_reality}).}

\changes{To obtain the zero-shot transfer of the policy in the real world, we train D-VAT with a more complex dynamic model, obtained by augmenting system \eqref{sysmodel} with the angular velocity, the thrust dynamics and the effect of air drag: 
\begin{equation}\label{vmodel}
\begin{bmatrix}
\ddot{p}\\ 
\dot{R}\\ 
\dot{\omega}\\ 
\dot{f}
\end{bmatrix} = 
\begin{bmatrix}
\frac{1}{m}(R_3f+f_{drag}) - g\\ 
R\left [  \omega\right ]_{\times}\\ 
J^{-1}(k_{\omega}(\omega_{cmd}-\omega)-\left [ \omega \right ]_\times  J \omega )\\
k_f(f_{cmd}-f)
\end{bmatrix} ,
\end{equation}
where $J$ is the inertia matrix, $f_{cmd}$ and $\omega_{cmd}$ are the commanded total thrust and body rates provided by D-VAT, $k_{f}$ and $k_{\omega}$ are suitable scalar gains, and $f_{drag} = -K_v\dot{p}$ is a linear drag term, being $K_v$ the drag coefficient matrix. The following parameter values have been employed: 
$J=\text{diag}(0.0030, 0.0045, 0.0028)$ kgm$^2$ and  $K_v=\text{diag}(0.3, 0.3, 0.15)$. Moreover, we set ${k_f=20}$ and $k_\omega=0.06$, resulting in a thrust settling time and control torque compatible with the MAV actuator specifications.
To improve the robustness with respect to model uncertainties we follow the strategy in \cite{dionigi2023exploring} and on each episode of the training phase we randomize the values of $m$, $J$, $g$ and $K_v$ up to $\pm 10\%$ of their nominal values, and add random delays up to a maximum settling time for the actuators of $0.25s$}.

\changes{To assess the performance on both indoor and outdoor environments, the experiments are performed on the Office and the Park visual scenarios. In each scenario, we consider four different trajectories for the target: a planar eight-shape, a planar rectangle, a 3D eight-shape and a 3D spiral.
The performance in Table \ref{tab:real} and the qualitative results in the attached video show that D-VAT achieves remarkable tracking performance even when deployed on a real platform without fine-tuning. Our approach achieves results comparable to those obtained in simulation (see Tables \ref{tab:results} and \ref{tab:velocity}), proving its generalization capabilities against unmodeled dynamics.}


\begin{table}[t]
\renewcommand{\arraystretch}{1.3}
\centering
\caption{\changes{Experimental results with Mixed Reality}\vspace{-1.0em}}
\resizebox{\columnwidth}{!}{
\label{tab:real}
\changes{\begin{tabular}{|c|c|c|c|c|c|c|c|c|}
\hline
\multirow{3}{*}{\shortstack{Target\\ Trajectory}} & \multicolumn{8}{c|}{Experimental Scenarios and Metrics}\\
\cline{2-9} & \multicolumn{4}{c|}{Outdoor (Park)} & \multicolumn{4}{c|}{Indoor (Office)}\\
\cline{2-9}
& $P_{\theta}$ & $P_{\varphi}$ & $P_{\rho}$ & $P_{c}$ & $P_{\theta}$ & $P_{\varphi}$ & $P_{\rho}$ & $P_{c}$\\
\hline
2D Eight-shape & $  0.93$ & $  0.94$ & $  0.72$ & $  0.86$ & $  0.95$ & $  0.94$  & $  0.69$ & $  0.86$\\
\hline
3D Eight-shape & $  0.93$ & $  0.94$ & $  0.71$ & $  0.86$ & $  0.94$ & $  0.94$  & $  0.67$ & $  0.85$\\
\hline
2D Rectangular-shape & $  0.94$ & $  0.94$ & $  0.76$ & $  0.88$ & $  0.93$ & $  0.94$  & $  0.67$ & $  0.85$\\
\hline
3D Spiral-shape & $  0.92$ & $  0.92$ & $  0.74$ & $  0.86$ & $  0.91$ & $  0.89$  & $  0.68$ & $  0.82$\\
\hline
\end{tabular}}}
 \vspace{-1.8em}
\end{table}

\section{Conclusions} \label{sec:conclusions}

In this work, we proposed D-VAT, an end-to-end visual active tracking approach for MAV systems. The D-VAT agent is trained by exploiting an asymmetric actor-critic DRL formulation. \changes{D-VAT computes thrust and angular velocity commands for the tracker MAV directly from input images. Experiments against different baselines show that our approach achieves a superior tracking performance and it is capable of generalizing over scenarios that considerably differ from those used during training, including real  platforms.}

Currently, D-VAT can track vehicles whose appearance is similar to that of the target MAV used for the optimization. Future work will consider methodologies to make the tracker agent independent from the appearance of the target \changes{MAV}.

\balance
\bibliographystyle{IEEEtran}
\bibliography{bibliografia}

\end{document}